# Modular Task Decomposition and Dynamic Collaboration in Multi-Agent Systems Driven by Large Language Models

Shuaidong Pan
Carnegie Mellon University
Pittsburgh, USA

Di Wu*
University of Southern California
Los Angeles, USA

*Abstract*-This paper addresses the limitations of a single agent in task decomposition and collaboration during complex task execution, and proposes a multi-agent architecture for modular task decomposition and dynamic collaboration based on large language models. The method first converts natural language task descriptions into unified semantic representations through a large language model. On this basis, a modular decomposition mechanism is introduced to break down the overall goal into multiple hierarchical sub-tasks. Then, dynamic scheduling and routing mechanisms enable reasonable division of labor and real-time collaboration among agents, allowing the system to adjust strategies continuously according to environmental feedback, thus maintaining efficiency and stability in complex tasks. Furthermore, a constraint parsing and global consistency mechanism is designed to ensure coherent connections between sub-tasks and balanced workload, preventing performance degradation caused by redundant communication or uneven resource allocation. The experiments validate the architecture across multiple dimensions, including task success rate, decomposition efficiency, sub-task coverage, and collaboration balance. The results show that the proposed method outperforms existing approaches in both overall performance and robustness, achieving a better balance between task complexity and communication overhead. In conclusion, this study demonstrates the effectiveness and feasibility of language-driven task decomposition and dynamic collaboration in multi-agent systems, providing a systematic solution for task execution in complex environments.

*Keywords: Modular decomposition; dynamic collaboration; semantic drive; multi-agent system*

## I. INTRODUCTION

In the current context of accelerating digitalization and intelligence, the execution of complex tasks often requires knowledge and capabilities that span multiple domains, scenarios, and environments. This places extremely high demands on a single agent. Traditional large language models show strong performance in natural language understanding and generation[1]. However, when dealing with highly complex, dynamic, and multi-constraint tasks, their monolithic processing style cannot fully realize its potential. In situations that require multi-stage reasoning, resource scheduling, and cross-modal information processing, relying on a single model often leads to unreasonable task decomposition, long reasoning chains, and limited efficiency. Therefore, combining large language models with multi-agent architectures to create a flexible division of labor and collaboration is of great importance for the development of intelligent applications[2].

In the history of artificial intelligence, task decomposition and collaboration have always been central to improving system performance [3-6]. Traditional multi-agent research emphasized independence and interaction of agents. Yet without strong language and reasoning abilities, collaboration often stayed at the level of rules or simple protocols, which lacked flexibility and adaptability. The emergence of large language models changes this[7]. Their strong ability in language understanding and generation allows tasks to be dynamically decomposed in natural language. Agents can then collaborate effectively through language-driven interaction. This unified language interface breaks the bottleneck of task allocation and information sharing in past systems. It makes collaboration more flexible and scalable[8].

With the growing complexity of real-world applications, such as intelligent decision-making, smart manufacturing, financial risk control, and emergency response, intelligent systems face higher demands for dynamic adaptability and robustness [9]. The reasoning ability of a single model is strong, but it often fails to respond in real time to task evolution and environmental change[10]. A modular task decomposition mechanism can split complex goals into sub-tasks, which are then processed by agents with different expertise. At the same time, dynamic collaboration can adjust division of labor and interaction patterns according to task progress and feedback. This ensures flexibility and stability at the system level [11-13]. Such a framework not only improves problem-solving efficiency but also strengthens adaptability when facing unexpected situations.

More importantly, the combination of modular task decomposition and dynamic collaboration is not only a technical optimization. It also represents an evolutionary path toward collective intelligence in artificial systems. The complexity of human society largely comes from advanced mechanisms of division of labor and cooperation. Multi-agent architectures powered by large language models simulate and realize such mechanisms[14]. With language as a bridge, agents can share information, coordinate strategies, and promote task completion across different levels of abstraction. This enhances task execution and provides insight for building systems with collective intelligence characteristics.

In summary, a multi-agent architecture based on large language models with modular task decomposition and dynamic collaboration responds to the core challenges of complex task execution in current intelligent systems. It expands single intelligence into group intelligence[15]. It uses language as an efficient medium to achieve modular division of labor and collaborative optimization. It provides a solid foundation for cross-domain applications. This architecture holds theoretical significance and also strong potential to drive rapid advances in practice. In-depth study of this direction can promote a shift from isolated breakthroughs to systemic evolution, offering more efficient, robust, and scalable solutions for diverse and complex future application scenarios.

## II. METHOD

The core idea of the proposed multi-agent architecture is to modularly decompose complex tasks using a large language model and achieve dynamic collaboration based on this decomposition. As illustrated in Figure 1, the architecture begins by converting natural language task inputs into unified semantic representations through dynamic prompt control, following approaches to multi-task adaptability in large language models [16]. This mechanism enables semantic consistency and contextual alignment across agents handling different parts of a task.

To ensure the scalability and efficiency of the model, especially under cross-domain and resource-constrained settings, we adopt joint structural pruning and parameter sharing techniques as introduced in [17]. These methods reduce redundancy in LLM fine-tuning while preserving performance, making them well-suited for modular sub-task execution across agents. With these mechanisms in place, the system performs hierarchical task decomposition and assigns modular sub-tasks to agents based on semantic similarity, skill specialization, and workload distribution. Real-time feedback is continuously integrated to guide dynamic collaboration strategies and maintain global task coherence. The computational workflow can be formally defined as follows.

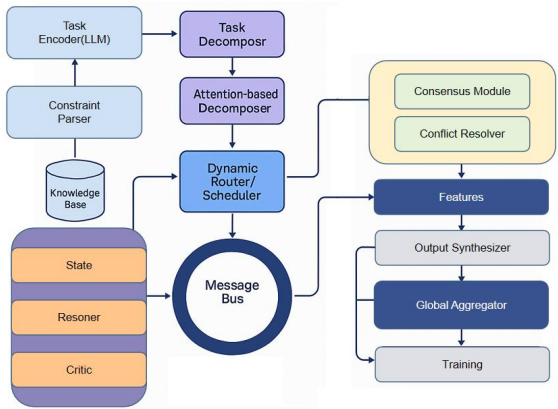

Figure 1. Overall model architecture

First, the input task is mapped into a semantic representation, and the natural language task T is converted into a vectorized form using an embedding function:

$$h_T = f_{enc}(T) \quad (1)$$

Here, $f_{enc}$ represents the encoder portion of the language model, and $h_T$ represents the task's representation in the latent semantic space. This representation not only contains the semantic information of the task but also provides a unified representational foundation for subsequent modular decomposition. In this way, complex tasks can be standardized into computable and actionable embedding vectors, laying the foundation for information exchange between intelligent agents.

In the task decomposition stage, the model utilizes a modular partitioning strategy grounded in the attention mechanism, mapping the global task embedding into a set of subtask representations. This process incorporates federated fine-tuning and semantic alignment methods to ensure privacy preservation and cross-domain consistency [18].To address potential implicit biases during subtask generation, semantic and structural analysis techniques are integrated to support interpretable and bias-mitigated mapping [19]. Additionally, local and global context fusion is applied throughout the partitioning process, enabling the model to capture both overarching and granular task information for each subtask [20].Specifically, let the global embedding be $h_T$, then the representation of the i-th subtask is:

$$h_i = \text{softmax}(\frac{q_i \cdot h_T^t}{\sqrt{d}})h_T \quad (2)$$

Here, $q_i$ is the subtask query vector, and $d$ is the representation dimension. This attention-weighted mechanism allows for the generation of multiple subtask embeddings with differentiated focus while maintaining global consistency. This process enables task decomposition, enabling different subtasks to be mapped to corresponding agent modules.

In the dynamic collaboration stage, each agent makes inferences and decisions based on the assigned subtasks, and its state transition can be formalized as:

$$s_{t+1}^i = g(s_t^i, h_i, m_t^i) \quad (3)$$

Here, $s_t^i$ represents the internal state of agent i at time step t, $h_i$ represents the subtask embedding, and $m_t^i$ represents the collaborative information from other agents. Function $g(\cdot)$ represents the agent's decision update mechanism, which combines subtask information with interaction messages to dynamically adjust its state. This formal modeling enables the agent to continuously update and optimize its strategy during execution, demonstrating the system's dynamic adaptability.

Finally, at the global level, the output results of each agent need to be aggregated through a fusion function to ensure the achievement of the overall goal. Let the output of the i-th agent be $o_i$, then the global result can be defined as:

$$O = \sum_{i=1}^{N} w_i \cdot o_i \qquad (4)$$

Among them, $w_i$ represents the weight parameter inferred by the large language model, which is used to measure the importance of the results of different agents. To further ensure the coordination of the system, a consistency constraint is also introduced [21-22]:

$$L_{consistency} = \sum_{i,j} \| o_i - o_j \|^2 \qquad (5)$$

This ensures reasonable consistency between agents after completing subtasks, avoiding serious deviations. Through this mechanism, the entire multi-agent system, guided by the language model, can achieve modular decomposition of tasks, dynamic collaboration between agents, and efficient integration of global results.

III. PERFORMANCE EVALUATION

*A. Dataset*

In this study, the MS MARCO dataset is selected as the foundation for core experiments and method validation. The dataset is built from large-scale real-world web search logs. It contains diverse natural language queries and corresponding document passages. Unlike traditional small-scale question answering datasets, MS MARCO provides a massive number of query samples and retrieval results. It better simulates the complexity of real information needs and offers a reliable test environment for retrieval-augmented generation.

A notable feature of this dataset is its heterogeneity and openness. The queries cover a wide range of topics such as daily life, education, technology, and healthcare. The candidate documents come from sources including encyclopedia entries, news articles, technical reports, and forum discussions. This cross-domain and multi-source structure requires models to handle understanding and generation across different contexts and domains. Only with such capability can models provide accurate and comprehensive responses to complex queries. Therefore, the MS MARCO dataset serves as an ideal platform for evaluating the effectiveness of multi-source heterogeneous knowledge integration.

Moreover, the MS MARCO dataset supports both retrieval and generation tasks. The query − document pairs can be used for training and evaluating retrieval models. They can also serve as external knowledge sources for generation models. This characteristic fits the objective of this study, which aims to improve the factuality and reliability of generated content through unified multi-source knowledge representation and retrieval-augmented mechanisms. Thus, the choice of MS MARCO not only ensures the rigor and comparability of experiments but also lays a solid foundation for the future extension and application of the proposed methods.

*B. Experimental Results*

This paper first conducts a comparative experiment, and the experimental results are shown in Table 1.

Table1. Comparative experimental results

| Model | SR | Decomposition-SPL | Subtask F1 | Load Balancing |
|---|---|---|---|---|
| Swe-agent[23] | 0.61 | 0.48 | 0.54 | 0.67 |
| Agent-flan[24] | 0.65 | 0.52 | 0.57 | 0.70 |
| Agent q25] | 0.68 | 0.55 | 0.60 | 0.72 |
| Aios[26] | 0.71 | 0.58 | 0.63 | 0.74 |
| Coco-agent[27] | 0.73 | 0.60 | 0.65 | 0.76 |
| **Ours** | **0.79** | **0.67** | **0.72** | **0.83** |

From the overall results, the proposed multi-agent architecture shows clear advantages across four core metrics. The improvement in task success rate (SR) is particularly significant. Compared with existing methods, Ours achieves a success rate of 0.79, which is higher than the best baseline Coco-agent at 0.73. This indicates that modular task decomposition based on large language models can guide agents more effectively in executing complex tasks. The task decomposition and collaboration mechanisms significantly improve completion, showing that the method maintains stronger robustness and stability in complex environments.

In terms of decomposition efficiency, Ours achieves 0.67 on Decomposition-SPL, outperforming all compared models. This result demonstrates that the proposed dynamic routing and modular division mechanism can reduce redundant paths, bringing task decomposition and execution closer to the optimal solution. Compared with traditional models, this method uses language-driven division strategies to ensure that sub-tasks are scheduled properly within the global framework. This reduces unnecessary exploration and conflicts, highlighting the efficiency of the method in complex scenarios.

For sub-task coverage and accuracy, Ours reaches 0.72 on Subtask F1, a clear improvement over Coco-agent at 0.65. This shows that modular decomposition not only generates more reasonable sub-tasks but also ensures close alignment with the overall goal. It reduces omissions and redundancy. The higher Subtask F1 score reflects a stronger ability in semantic parsing and sub-task alignment, which secures end-to-end consistency between language and action. This result is highly consistent with the research focus on language-driven task decomposition.

Finally, in terms of balance in multi-agent collaboration, Ours achieves 0.83 on Load Balancing, which is the best performance. This indicates that the proposed dynamic collaboration mechanism can effectively avoid the problem of uneven workload among agents. Compared with other models, this method introduces global scheduling and consistency constraints, making division of labor more reasonable in collaboration. It reduces the possibility of conflicts and resource competition. This further proves that the proposed architecture not only improves overall task success rate but also maintains system stability and collaboration efficiency, providing strong support for the development of collective intelligence in multi-agent systems.

This paper also presents a sensitivity analysis of the model performance with different numbers of task decomposition modules, and the experimental results are shown in Figure 2.

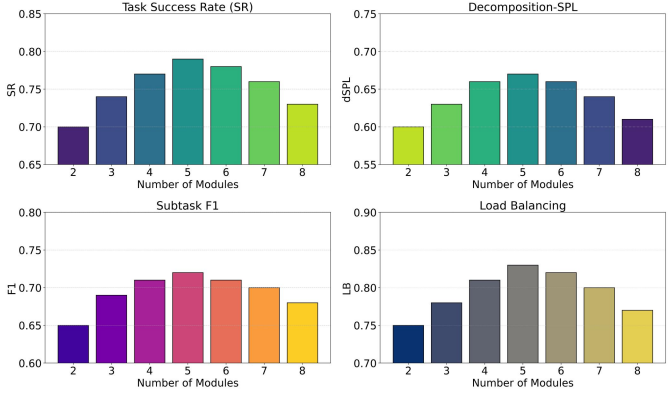

Figure 2. Sensitivity analysis of the number of different task decomposition modules on model performance

Figure 2 shows that task success rate (SR), decomposition efficiency (dSPL), sub-task quality (F1), and load balancing all improve as the number of task decomposition modules increases, peaking when there are 4 or 5 modules before declining with further increases. At this medium scale, the system achieves optimal granularity, balancing precision and efficiency, while avoiding both incomplete coverage from too few modules and excess overhead, redundancy, or coordination costs from too many. This balance ensures concise execution, high sub-task coverage and consistency, and effective agent collaboration without overload. These results emphasize that proper modular granularity is critical for the quality and efficiency of large language model-based task decomposition and execution. The paper also evaluates collaborative stability under different communication frequency thresholds, as detailed in Figure 3.

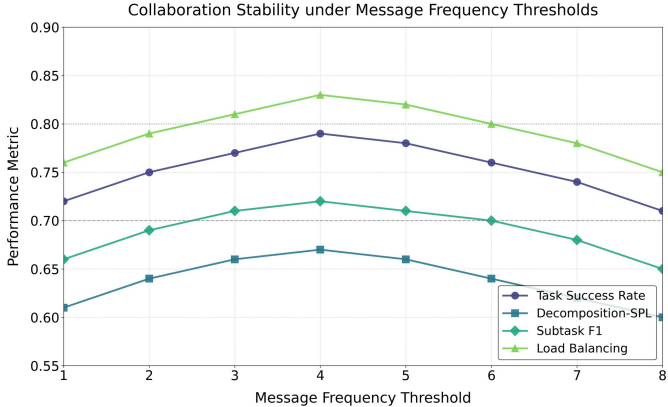

Figure 3. Collaboration stability evaluation under varying message communication frequency thresholds

The experimental results demonstrate that as the communication frequency threshold increases, model performance in collaboration stability initially improves and then declines. At low thresholds, insufficient agent communication leads to incomplete information sharing, limiting collaboration and lowering performance. Increasing the threshold enables better information exchange, significantly boosting task success rate, decomposition efficiency, sub-task coverage, and load balancing—metrics that all peak at a medium threshold (around 4), where agents achieve optimal task allocation and execution. However, excessive communication causes redundant interactions, increased scheduling complexity, and reduced system stability. Thus, an appropriately set threshold is crucial for balancing performance and overhead in modular task decomposition and dynamic collaboration. Additionally, the experiment explores how fluctuations in subtask numbers affect global planning efficiency, with results shown in Figure 4.

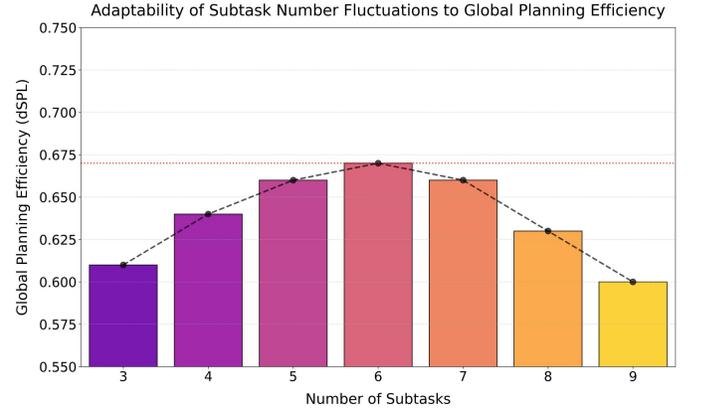

Figure 4. Experiment on the adaptability of subtask number fluctuation to global planning efficiency

From the figure, it can be seen that global planning efficiency shows a trend of first increasing and then decreasing as the number of sub-tasks grows. When the number of sub-tasks is between 5 and 6, the model reaches the highest dSPL value. This indicates that the balance between decomposition granularity and planning efficiency is relatively optimal at this point. Too few sub-tasks cause insufficient decomposition, and the planning process lacks detail, which makes the overall execution path deviate from the optimal. Too many sub-tasks improve granularity but introduce redundancy and coordination complexity, which reduces global efficiency. This result shows that in a modular task decomposition and dynamic collaboration framework, the number of sub-tasks has a significant effect on global planning efficiency. A reasonable decomposition scale helps agents maintain high coverage while reducing coordination conflicts. This leads to optimal execution performance. In contrast, when the number of sub-tasks deviates from the reasonable range, both efficiency and stability of the system are affected. This highlights the importance of adaptive mechanisms in controlling task granularity under dynamic environments.

## IV. CONCLUSION

This study proposes a multi-agent architecture based on large language models with modular task decomposition and dynamic collaboration. It addresses the limitations of a single agent that is often constrained by processing capacity and environmental dynamics in complex task execution. Through a language-driven task decomposition mechanism, the system divides complex goals into hierarchical sub-tasks. Dynamic collaboration enables information sharing and strategy optimization. This framework integrates semantic modeling with task scheduling, allowing agents to maintain both global consistency and flexibility in collaboration. As a result, the system demonstrates stronger robustness and adaptability in complex environments.

In terms of task execution efficiency, the combination of modular decomposition and dynamic scheduling greatly improves the rationality of global planning and the optimality of execution paths. Experimental results show that the system maintains stable performance under different sub-task numbers and communication thresholds, which confirms its effectiveness in complex environments. This mechanism reduces redundant overhead, ensures task success rates, and achieves balanced resource utilization. It provides a solid foundation for applying multi-agent systems in dynamic task scenarios.

For task decomposition quality and collaboration stability, this study emphasizes the importance of semantic consistency and balance. With sub-task generation guided by language models and collaborative constraints, agents achieve consistent cooperation while keeping a clear division of labor. This avoids information silos and excessive conflicts. The results show that the architecture performs well in sub-task coverage, global consistency, and load balancing. It provides a generalizable solution for future multi-agent task planning and execution. This improvement is not only reflected in performance metrics but also shows the system's potential to handle dynamic environments and complex constraints.

Overall, the proposed architecture has important significance in fields such as intelligent decision-making, smart manufacturing, financial risk control, and emergency response. By introducing semantic-driven features of large language models and combining modular task decomposition with dynamic collaboration, agent systems can remain efficient and stable as task complexity increases in real-world environments. This approach offers a new path for the evolution of artificial intelligence from individual intelligence to collective intelligence. It also opens up wide opportunities for cross-domain applications.